\documentclass[letterpaper, 10 pt, conference]{ieeeconf}  

\IEEEoverridecommandlockouts                              

\usepackage{graphics} 
\usepackage{epsfig} 
\usepackage{mathptmx} 
\usepackage{times} 
\usepackage{amsmath} 
\usepackage{amssymb}  
\usepackage{amsfonts}
\usepackage[caption=false]{subfig}
\usepackage{graphicx}
\usepackage[normalem]{ulem}

\usepackage[printonlyused,withpage,nolist,nohyperlinks]{acronym}
\begin{acronym}

\acro{ATE}{Absolute Trajectory Error}
\acro{ICP}{Iterative Closest Point}
\acro{SLAM}{Simultaneous Localisation and Mapping}
\acro{ESDF}{Euclidean Signed Distance Field}
\acro{TSDF}{Truncated Signed Distance Field}
\acro{RMSE}{Root Mean Squared Error}
\acro{CNN}{Convolutional Neural Network}
\acro{AR}{Augmented Reality}
\acro{VR}{Virtual Reality}
\acro{SFC}{Safe Flight Corridor}

\end{acronym}

\usepackage{multicol}
\usepackage[bookmarks=true]{hyperref} 
\hypersetup{colorlinks,breaklinks,
linkcolor=[rgb]{0.5,0.,0.},
citecolor=[rgb]{0.000,0.427,0.173},
urlcolor=[rgb]{0.031,0.318,0.612}}

\usepackage{booktabs}
\usepackage{multirow}
\usepackage{tabularx}
\usepackage{algorithm2e}
\SetAlgoLined
\SetKwProg{MyStruct}{Struct}{ contains}{end}

\newcommand\edit[1]{{\color{black}#1}}

\newcommand{\SEthree}{\ensuremath{\mathrm{SE}(3)}\xspace}

\title{\LARGE \bf
Multi-Resolution 3D Mapping with Explicit Free Space Representation
for Fast and Accurate Mobile Robot Motion Planning
}

\author{Nils Funk$^{1,2}$, Juan Tarrio$^{2}$, Sotiris Papatheodorou$^{1}$, Marija Popovi\'{c}$^{1}$, Pablo F. Alcantarilla$^{2}$, Stefan Leutenegger$^{1,2}$
\thanks{$^{1}$ Smart Robotics Lab, Imperial College London, UK {\tt\small \{nils.funk13, s.papatheodorou18, m.popovic, s.leutenegger\}@imperial.ac.uk}}
\thanks{$^{2}$ SLAMcore Ltd.,  London, UK {\tt\small \{first.name\}@slamcore.com}}
\thanks{The video is available at: \url{https://youtu.be/XlnO3GBNPvc}}
}

\begin{document}

\maketitle

\begin{abstract}
With the aim of bridging the gap between high quality reconstruction and robot motion planning, we propose an efficient system that leverages the concept of adaptive-resolution volumetric mapping, which naturally integrates with the hierarchical decomposition of space in an octree data structure. Instead of a Truncated Signed Distance Function (TSDF), we adopt mapping of occupancy probabilities in log-odds representation, which allows to represent both surfaces, as well as the entire free, i.e.\ observed space, as opposed to unobserved space. We introduce a method for choosing resolution -on the fly- in real-time by means of a multi-scale max-min pooling of the input depth image. The notion of explicit free space mapping paired with the spatial hierarchy in the data structure, as well as map resolution, allows for collision queries, as needed for robot motion planning, at unprecedented speed. We quantitatively evaluate mapping accuracy, memory, runtime performance, and planning performance showing improvements over the state of the art, particularly in cases requiring high resolution maps.
\end{abstract}


\section{Introduction} 
\label{sec:introduction}

The past years have brought impressive advancements in the field of dense environment mapping, fueled by the advent of RGB-D cameras and ever more powerful processors, including GPUs. Applications of (near-)real-time 3D dense mapping systems are vast, ranging from digital twins and Augmented/Virtual Reality to mobile robotics.
Recent technological advances are inciting the use of mobile robots for exploration and monitoring tasks. Their growing versatility and autonomy offer safe, cost-effective solutions in a wide range of applications, including aerial surveillance, infrastructure inspection, and search and rescue \cite{Nex2014}. However, to fully exploit their potential, a key task is enabling light-weight platforms to operate autonomously in unknown, unstructured environments with limited on-board computational resources. Map representations are thus required that can accommodate both high-fidelity reconstructions of the environment as well as perform fast online planning.

\begin{figure}[t]
    \centering
    \includegraphics[trim={0.1cm 0.1cm 0.1cm 0.9cm},clip, width=0.985\columnwidth]{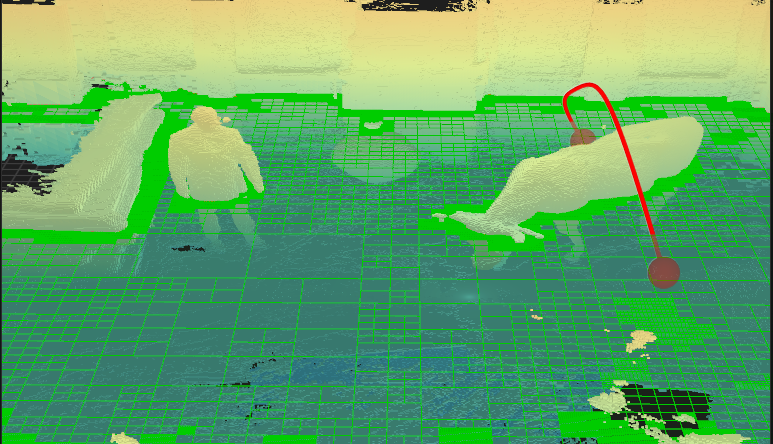}
    \caption{The main goal of our system is to provide a navigable 3D occupancy map with a defined notion of observed free space in small to large scale scenarios, using adaptive resolution to constrain memory consumption. The figure shows the output of our mapping pipeline for the \textit{Cow and Lady} RGB-D dataset \cite{Oleynikova2017} alongside a planned trajectory (red). The chosen voxel resolutions for free space encoding are shown for a map slice.}
    \label{fig:cover}
\end{figure}

There are several efficient mapping methods for motion planning that rely on 3D volumetric representations. A common strategy is to use probabilistic occupancy maps using octree structures for quick access \cite{Hornung2013,Vespa2018}. More recently, signed-distance-based frameworks \cite{Oleynikova2017,Han2019,Vespa2019} have become popular as they allow fast map query operations in online settings. The main drawback of previous approaches is that their computational performance degrades drastically with the discretisation of the environment, since they operate on map data in the same way regardless of the occupancy status and geometry of the underlying space. As a result, these methods are only suitable for scenarios requiring coarse reconstructions or navigating in areas with relatively large obstacles. In contrast, our work addresses the challenge of trading off mapping accuracy  against computational efficiency for planning in online, on-board robotic applications that require \textit{both}.

In this paper, we introduce a volumetric adaptive-resolution \textit{dense} mapping framework that supports multi-resolution queries and data integration using occupancy mapping \cite{Hornung2013,Elfes1989,Thrun2003}. Contrary to methods based on signed distance functions, we continuously maintain a high resolution 3D octree representation of observed  occupied and \textit{free} space in real-time. 

Our key insight is to recognise the lack of a concise method for defining required resolution in the context of occupancy mapping, with few approaches tackling this problem by extending single resolution approaches with ad-hoc heuristics \cite{Vespa2018}. 
As a result, we designed a novel integration algorithm that selects resolution by constraining the induced sampling error in the observed occupancy, splitting an octree in a coarse-to-fine fashion until a desired accuracy is reached. Central to this idea is the introduction of a multi-scale max-min pooling of the input depth image which enables real-time operation by providing a conservative indication of measured depth variation in any given volume with only a few queries.

To further enhance performance, we \edit{modified} the data structure of \cite{Vespa2019} which uses mip-mapped voxel blocks, and carefully designed a new scale selection and data propagation scheme between levels. This allowed us to increase computational and memory efficiency without introducing reconstruction artefacts.

While our method focuses on RGB-D mapping, it is flexible to different sensor modalities. This is shown by evaluating our system in a wide range of datasets, from synthetic to large scale LIDAR, showing significant improvements in reconstruction accuracy, runtime performance, and planning performance against state-of-the-art approaches.
In summary, the main contributions of this paper are:

\begin{enumerate}
    \item A dense volumetric multi-resolution system, \edit{comprising a data structure, fusion method and sensor models which together enable real-time online probabilistic occupancy mapping and accurate surface reconstruction, by consistently} representing free and occupied space to fine resolutions where needed; we will release our reference implementation open-source upon acceptance of this work.
    \item A novel fast map-to-image allocation \edit{algorithm}  and integration method, that seamlessly adapts the map to different scene scales and sensor modalities 
    \item A comprehensive evaluation with respect to the state-of-the-art revealing vast improvements in the trade-offs of mapping accuracy, speed, memory consumption, tracking accuracy, and planning performance.
\end{enumerate}

\section{Related Work}
\label{sec:related_work}

A large body of literature addresses the problem of obtaining a dense 3D representation of the world. In this section, we overview recent work, focusing on volumetric reconstruction methods suitable for real-time robotic applications running on constrained hardware. This leaves aside most batch integration methods, where a map is only available for planning at the end of the mapping run; note that most newer deep learning based methods fall in this category.

A major milestone in online RGB-D 3D reconstruction systems is KinectFusion \cite{Newcombe2011}, which enables dense volumetric modelling in real-time at sub-centimetre resolutions. However, it is limited to small, bounded environments, as the mapping is locked to a fixed volume with a pre-defined voxel resolution, and requires GPGPU processing to achieve real-time performance. To improve scalability, several extensions to the original algorithm have been proposed. One possibility is to use moving fixed-size sliding volumes \cite{Whelan2012,Usenko2017} to achieve mapping in a dynamically growing space. Another strategy is to exploit memory-efficient data structures, such as octree-based voxel grids \cite{Vespa2018,Zeng2013,Steinbrucker2014} or hash tables \cite{Niessner2013,Klingensmith2015}, for quicker spatial indexing. More recently, deep learning methods tackling this problem in an incremental fashion are also being introduced \cite{Weder2020}. Despite impressive progress, most 3D mapping research has targeted the application of surface reconstruction, which aim to produce a high-quality mesh/point-cloud of a scene. From a navigation perspective, the concept of observed free space is equally or more important than surface accuracy. Unfortunately, most of these methods model free space only near the surface boundaries and do not generally distinguish between free and unvisited areas in initially unknown environments. Our work provides an explicit distinction between the two as necessary for robotic planning, exploration, and collision avoidance.

In the context of navigation, \ac{ESDF}-based mapping methods are commonly used for motion planning tasks as they provide distance information for trajectory optimisation strategies. Recently, significant work has been done on incrementally building \ac{ESDF} maps for planning in 3D using aerial robots, including the \textit{voxblox} \cite{Oleynikova2017} and \textit{FIESTA} \cite{Han2019} frameworks, which construct \ac{ESDF} maps from \ac{TSDF} maps and occupancy maps, respectively. However, as these methods are designed for fast on-board collision checking, they rely on coarsely disretised environments with voxel grid resolutions on the order of $\sim 20$\,cm magnitude. In contrast, our approach is also motivated by applications like close-up inspection \cite{Stent2015}, which require detailed scene reconstructions.

An alternative representation for planning is the occupancy map \cite{Elfes1989,Thrun2003}. In 3D, OctoMap \cite{Hornung2013} is a popular framework that uses hierarchical octrees to track occupancy probabilities as sensor data is received. Similar to the original work of \cite{Elfes1989}, it uses an inverse sensor model that efficiently approximates the posterior using an additive log-odds update equation, which resembles the \ac{TSDF} update procedure \cite{Newcombe2011}. 

However, while OctoMap works well with sparse LIDAR data, its performance degrades significantly as map resolution increases, as well as with noisier sensors. Interestingly, a very recent contribution shows improvements over OctoMap in terms of memory and run time by adaptively downsampling the pointcloud and integrating free space at lower resolutions  \cite{Duberg2020}, highlighting the importance of this topic. While this method uses a set of distance-based rules to decide the integration resolution (similar to \cite{Vespa2018}), leading in the fastest setups to non-conservative assignments of free space, our work tackles this problem by rigorously assessing the probabilistic (inverse) measurement model, introducing an approach for choosing a sampling resolution that ensures these errors do not happen, while also improving run time performance.    

Closest to our work, \cite{Vespa2018} proposed an efficient pipeline with an octree-based implementation for reconstruction and planning. Their approach supports either \ac{TSDF}-based or occupancy mapping using the spline inverse sensor model of \cite{Loop2016}. However, the use of multi-resolution is very basic and multiple assumptions limit its applicability to planning.
 
Subsequent work \cite{Vespa2019} extended this system to handle data integration with varying levels of detail and rendering at multiple resolution scales using \ac{TSDF} maps, but limited to surface reconstruction. Our method draws inspiration from this approach in terms of data structure and propagation; however, our focus is on volumetric occupancy-based representations for planning and space understanding, where the goal is to probabilistically classify all observed space into occupied, free, or unknown at high resolutions, while also providing a high quality (surface) reconstruction of the environment.

\begin{figure}[htb]
    \centering
    \includegraphics[width=\columnwidth]{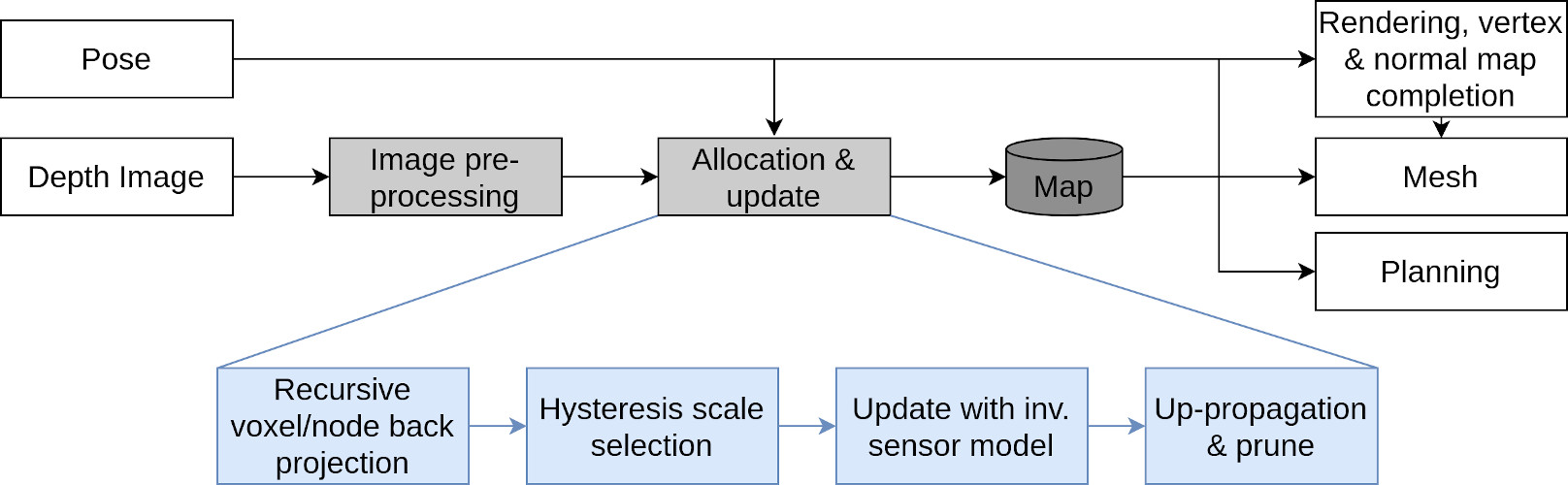}
    \caption{Overview of our system: The different stages of the mapping pipeline are shown in grey, while the blue boxes show the steps in the \textit{allocation and updating} procedure. The flow of information is illustrated by the arrows. The allocation and updating stages modify the map, while the rendering and planning stages utilise the map information.}
    \label{fig:block_diagram}
\end{figure}

\begin{figure}[htb]
    \centering
    \includegraphics[width=\columnwidth]{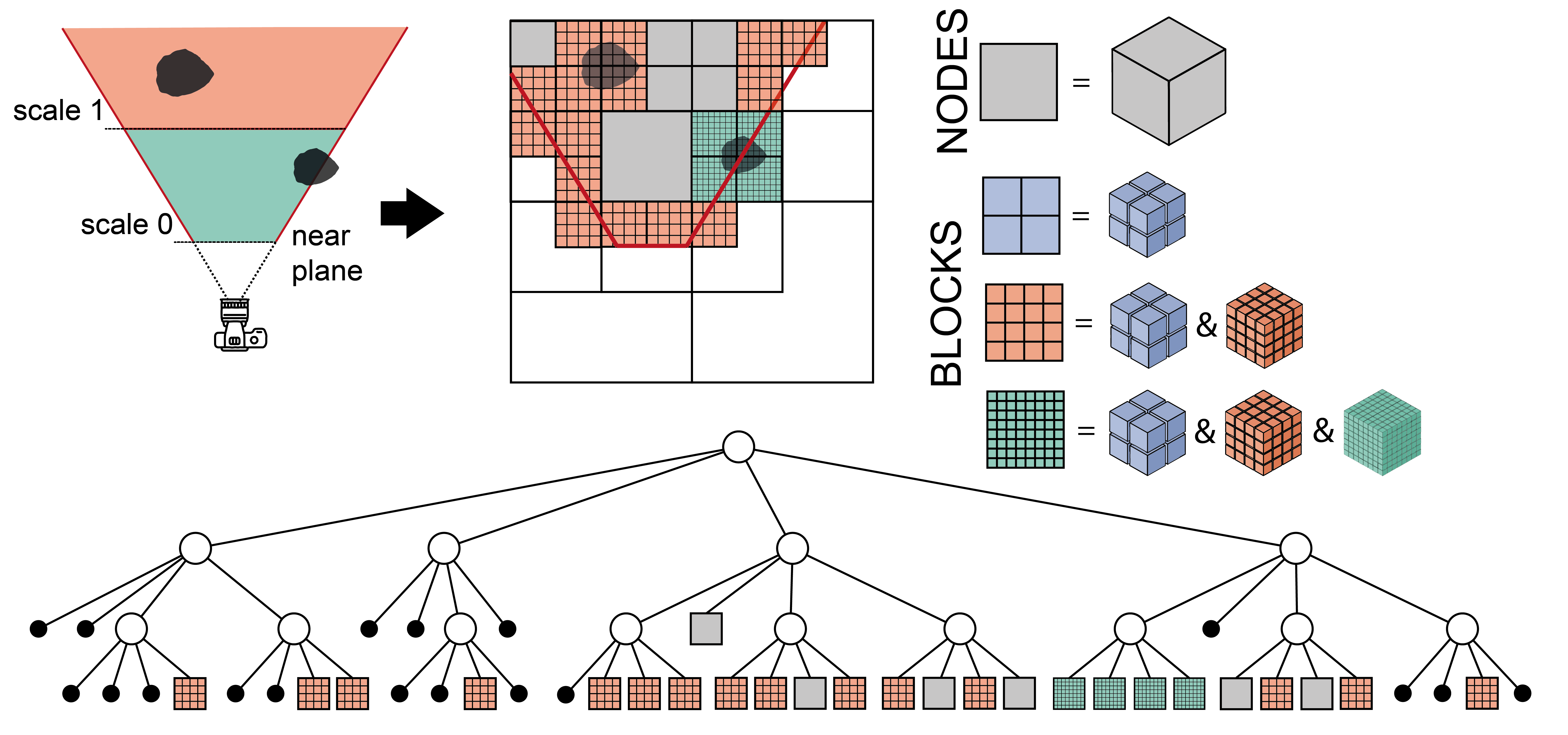}
    \caption{\edit{Overview of the data structure for a hypothetical 2D case. Data is represented in a two tier octree having voxel blocks in the lower levels. Depending on the distance based integration scale $s_c$ voxel blocks containing data up to different resolutions are allocated, resulting in important memory and time savings for cases where higher resolution is not needed. Note that in the given example the integration scale $s_c$ for blocks only containing free and unknown space has a lower bound of $s_c > 1$} (Section \ref{section:Fusion}).}
    \label{fig:quadtree}
\end{figure} 

\section{Multi-Resolution Occupancy Mapping}
\label{sec:mapping_approach}

Our library takes as input depth information and poses, incrementally computing a map which can be queried at any time for path planning, meshing, and rendering (see Fig. ~\ref{fig:block_diagram}). We also provide an \ac{ICP} module that can be switched on to obtain a full \ac{SLAM} system in the spirit of \cite{Newcombe2011}, as shown in Section~\ref{subsec:tracking}.

\edit{In order to represent an occupancy efficiently using adaptive resolution we use an octree where, similar to \cite{Vespa2018}, the last levels consist of densely allocated voxel blocks aggregating up to $8 \times 8 \times 8$ voxels at the finest scale. This two tier memory structure leverages flexibility in choosing resolution at the higher (octree) levels with efficient access at lower (voxel block) level. Taking inspiration from \cite{Vespa2019}, each voxel block stores a pyramid representation of itself enabling fast occupancy updates at different resolutions. However, unlike \cite{Vespa2019}, we save memory by not keeping all mipmap levels and only allocating data in each block down to a single dynamically changing integration scale $s_c$, in other words our design has different voxel block types which are changed dynamically depending on the needed integration scale. More importantly, we augment this method by also allowing data to be stored at node level when the voxel block resolution is finer than needed. A representation of this structure can be seen in Fig.~\ref{fig:quadtree} for a hypothetical 2D case. }

For a given point, the relevant occupancy information is maintained only at the lowest allocated voxel that contains it. Similar to \cite{Hornung2013}, upper non-leaf nodes store a max pooling of the children occupancy, enabling fast conservative queries of any given size. In addition to the max occupancy, we keep a Boolean state indicating whether unknown data is present in the children. This allows us to disambiguate a node as being partially or fully unobserved. In our system this pooling, or data up-propagation, is computed in each integration step, making it available for online planning at any given time. 

\subsection{Notation}
\label{sec:notation}
We denote $n$-dimensional vectors with lower-case, bold letters, e.g.\ $\mathbf{x} \in \mathbb{R}^n$.  Also, we denote the coordinate frame in which vectors are expressed with left subscripts, e.g.\ $_C\mathbf{x}$. We employ a World frame $\{W\}$ and the Camera frame $\{C\}$.
Euclidean transformations from coordinate frame $\{C\}$ to $\{W\}$ are denoted as $\mathbf{T}_{WC} \in \SEthree$.
We denote matrices with upper-case bold letters, e.g.\ $\mathbf{D} \in \mathbb{R}^{n \times m}$.

\subsection{Occupancy Map Fusion}
\label{subsec:occupancy_map}

Given a depth image $\mathbf{D}_k \in \mathbb{R}^{H \times W}$ and camera pose $\mathbf{T}_{WC_k}$ at time step $k$,
the probability that a point $_W\mathbf{p}$ in 3D space is occupied is assumed to depend on the distance to the camera and the corresponding depth measurement along the ray from the camera centre $_W\mathbf{c}$ to $_W\mathbf{p}$.
Given the depth measurement $z$ from the projection of point $_W\mathbf{p}$ into the camera, we represent the occupancy probabilities of one depth image  in 3D:
\begin{align}
    P_\mathrm{occ}\big(_W\mathbf{p} \, | \, \mathbf{D}_k, \mathbf{T}_{WC_k}\big)
    = P_\mathrm{occ} \big(_W\mathbf{p} \, | \, z
    = \mathbf{D}_k \big[\pi(\mathbf{T}_{WC_k}^{-1} {_W\mathbf{p}})\big] \big) \,, 
    \label{ec:p_occ_p}
\end{align}

which corresponds to the inverse sensor model \cite{Thrun2003}, a function of the depth along the ray. For brevity, it is referred to as $P_{\mathrm{occ}}(_W\mathbf{p}|z)$ in the following descriptions.
An alternative way of representing the occupancy probability is using log-odds, which allows for Bayesian updates to be additive as:
\begin{align}
    l_k (_W\mathbf{p}) &= \log \frac{P_{\mathrm{occ}}(_W\mathbf{p}|z)}{1 - P_{\mathrm{occ}}(_W\mathbf{p}|z)} \, , 
    \label{ec:log_odds_updt_og1} \\
    L_{k} (_W\mathbf{p}) &= L_{k-1} (_W\mathbf{p}) + l_k (_W\mathbf{p}) \, .
    \label{ec:log_odds_updt_og2}
\end{align}

In contrast to previous work, we do not accumulate the log-odds in a single sum, but instead use a weighted mean $\bar{L}_{k}(_W\mathbf{p})$, with weight $w_k$, similar to how it is done in \cite{Newcombe2011} \edit{providing additional information about the number of integrations}. Thus the mean log-odds is updated as:
\begin{align}
    \bar{L}_{k}(_W\mathbf{p}) &= \frac{\bar{L}_{k-1}(_W\mathbf{p}) \ w_{k-1} + l_{k}(_W\mathbf{p})}{w_{k-1} + 1} \, , 
    \label{eq:update_mean_log}\\
    w_k &= \min\{w_{k-1} + 1, w_{\max}\},
\end{align}
while the accumulated log-odds can be preserved:
\begin{align}
        L_{k}(_W\mathbf{p}) &= \bar{L}_{k}(_W\mathbf{p}) \ w_k \,.
    \label{ec:log_odds_updt}
\end{align}
By clamping the weight $w_k$ below a threshold $w_{\max}$,
the influence of outliers and dynamic objects is mitigated.  \edit{Moreover, using a maximum weight $w_{\max}$ rather than an occupancy threshold prevents the occupancy field at the surface from converging to a step function and does not require the use of a time based windowed updating step as in \cite{Vespa2018}}. 

\subsection{Inverse Sensor Model}
\begin{figure}[htb!]
    \centering
    \includegraphics[ width=0.98\columnwidth]{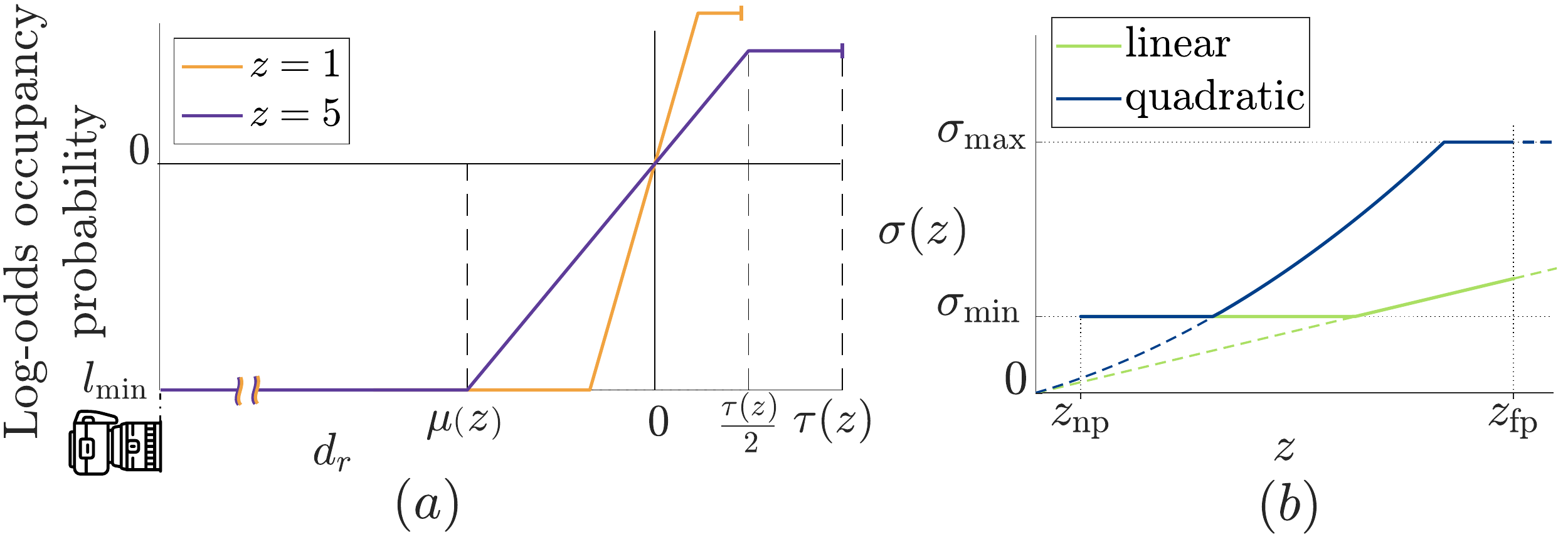}
    \caption{(a) Inverse sensor model for two measurements (1m, 5m) expressed as a function of the difference  $d_r$ from a query point  to the measured surface along the ray. Log-odd values in front of the surface are clipped at $l_{\min}$ reached at $\mu = 3 \sigma$ and grow linearly up to half the surface thickness $\tau(z)$. 
    (b) Distance-dependent growth of two sensor uncertainty models (linear and quadratic) within the minimum and maximum sensor range $z_{\text{np}}$ and $z_{\text{fp}}$ and sensor uncertainty $\sigma_{\min}$ and $\sigma_{\max}$.}
    \label{fig:sensor-model}
\end{figure}

For fast computations, our inverse sensor model is a piece-wise linear function based on \cite{Loop2016} and illustrated in Fig. \ref{fig:sensor-model} (a) operating directly in log-odds space.
Similarly to \cite{Loop2016}, we use a model where the measured surface position matches a log-odds value of zero.

We model  depth uncertainty $\sigma$ as a function of measured depth $z$, assuming it to be linearly or quadratically growing depending on the sensor type (see Fig. \ref{fig:sensor-model} (b)): quadratic for RGB-D and linear for LIDAR or synthetic (perfect) depth cameras.  
We relax the assumption of quadratic relation made in \cite{Loop2016} between the surface thickness $\tau(z) \propto z^2$  with a linear model $\tau(z) \propto z$ bounded by a minimum and maximum surface thickness $\tau_{\min}$ and $\tau_{\max}$, to avoid overgrowing of distant objects.

\subsection{Adaptive-Resolution Volume Allocation}
\label{subsec:map_to_cam}
 
The data structure is interpreted as a non-uniform partition of a continuous occupancy 3D scalar field \cite{Vespa2018}. Consequently we do not assign any probabilistic meaning to the size of the voxel storing a particular value and instead aim at choosing a partition of the tree which can represent the underlying continuous function up to a certain accuracy.

\begin{figure}[htb]
    \centering
    \includegraphics[trim={1.7cm 0 9.7cm 0},clip, width=0.49\columnwidth]{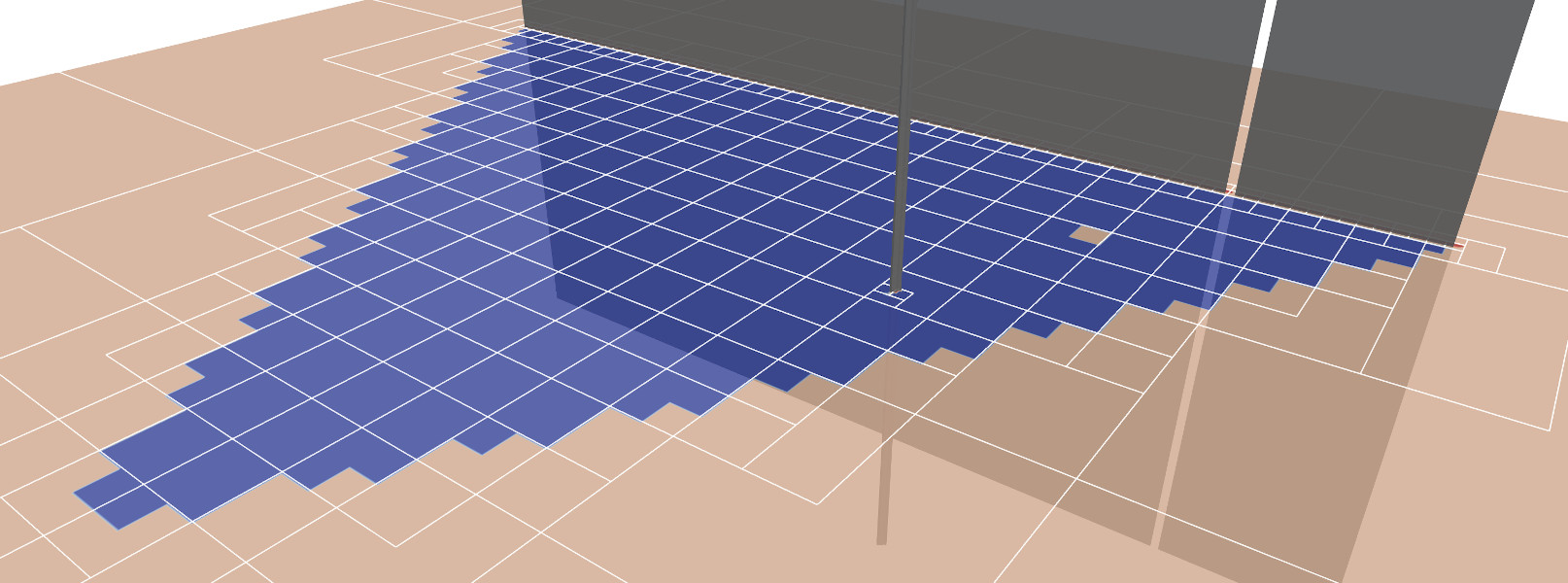}
    \includegraphics[trim={1.7cm 0 9.7cm 0},clip, width=0.49\columnwidth]{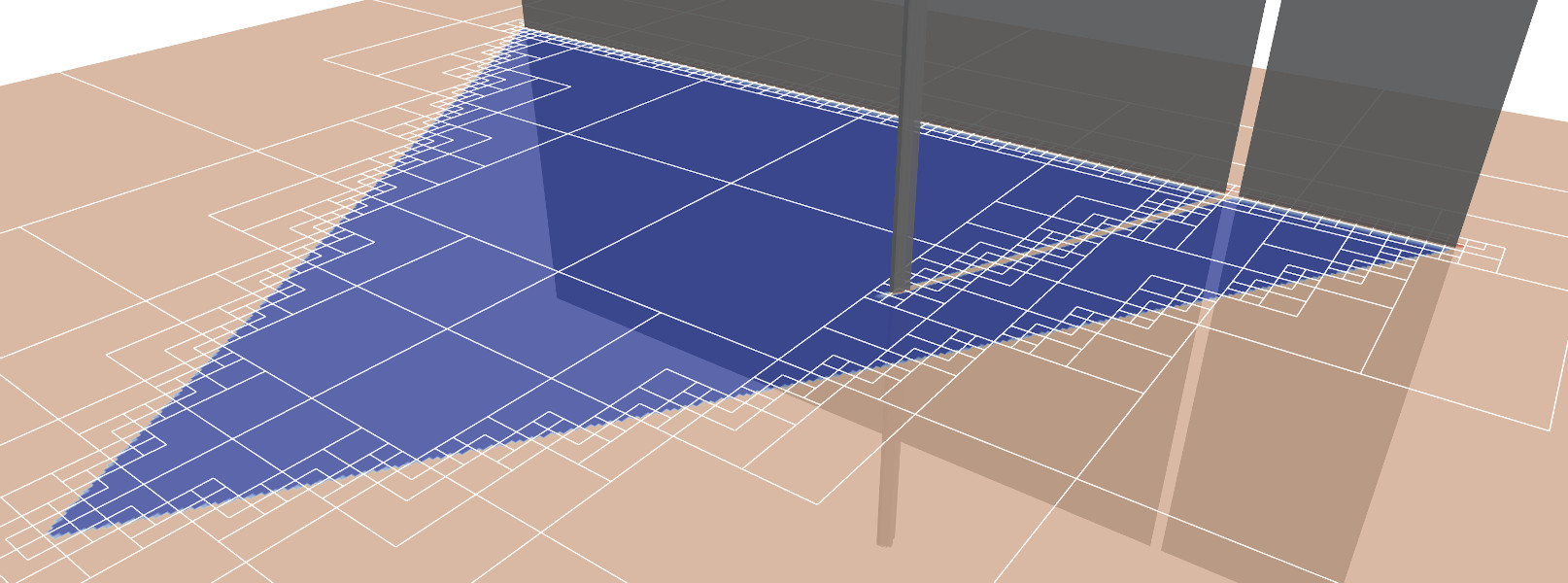}
    \caption{\edit{Comparison of free space volume allocation in OFusion \cite{Vespa2018} (\textbf{left}) and our new adaptive-resolution strategy (\textbf{right}) in an environment with a wall and vertical pole. Blue corresponds to free space and white lines indicate allocated voxels on a map slice. While OFusion ignores occupancy variations inside a voxel's volume, leading to erroneous assumptions of free space behind the pole, our method naturally integrates the volume at the appropriate scale, allocating obstacles and boundaries to unknown space at fine resolution, while representing free space at the coarsest possible level.}}
    \label{fig:motivation}
\end{figure}

Choosing this partition -on the fly- in a volumetric space is not a trivial task. Methods like \textit{OctoMap} \cite{Hornung2013} use a ray-casting scheme to allocate densely the observed space, simplifying resolution in a later stage called tree pruning. Newer methods like \textit{OFusion} \cite{Vespa2018} and \textit{UFOMap} \cite{Duberg2020} mitigate the need for dense allocation by using a set of heuristics to choose the needed resolution during the allocation step. The main problem with these approaches is illustrated in Fig.~\ref{fig:motivation}. The fact that resolution is chosen as part of the ray-casting and mainly as function of distance tends to ignore changes in occupancy that happen because of depth changes between pixels, which often require high resolution allocation to be accurately captured, leading to parts of space being erroneously labeled as free. This is particularly important in occlusions caused by thin objects and frustum boundaries. 

To solve this problem we take a radically different approach, discarding the raycasting and using a so called  `map-to-camera' allocation and updating process, thereby following an `as coarse as possible, as fine as required' mapping scheme.
Given a new depth image  $\mathbf{D}_k$, we start from the root of the octree analysing each node recursively in order to decide whether the variation of occupancy log-odds inside the node meets a bounding criterion:
\begin{equation}
    \max{|l_k(_W\mathbf{p}_i) - l_k(_W\mathbf{p}_j)|} < \epsilon
    \label{eq:split_crit}
\end{equation}
for every $_W\mathbf{p}_i, \, _W\mathbf{p}_j$ in the node volume. This criterion, which has been used for 3D data compression on an octree \cite{Knoll2006}, \edit{bounds the occupancy error from a sampling perspective, a desirable feature for safe navigation maps.}
If it is met, we update the node at the given scale;
otherwise, it is split into its eight children, and the process is recursively repeated until \eqref{eq:split_crit} is satisfied or voxel block level is reached. 

\begin{figure}[t]
    \centering
    \includegraphics[ width=1.00\columnwidth]{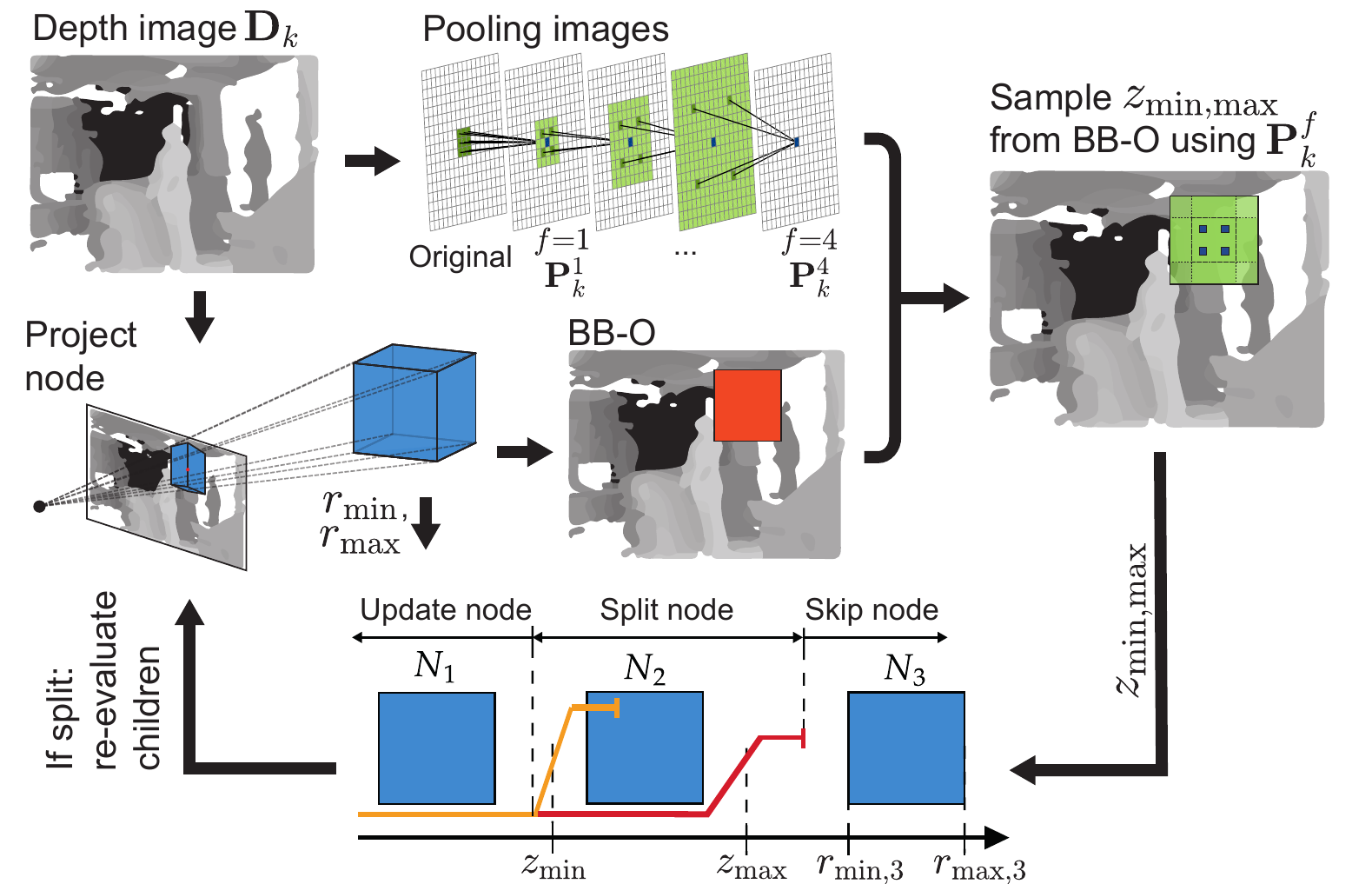}
    \caption{\edit{The process to decide whether a node should be updated, split or skipped. From the projection bounding box (BB-O), simple queries $\mathbf{P}^f_k[u,v]$ to the pooling image at the correct scale $f$ provide an span of measurements $[z_{\min}, z_{\max}]$  which is used alongside the inverse sensor model and the span of query depths $[r_{\min}, r_{\max}]$ in the node's volume to make a decision (shown for 3 possible cases $N_1-N_3$). The node is also split if pooling indicates it projects into partly unknown data or crosses the frustum.}}
    \label{fig:voxels_3d}
\end{figure}

\edit{
Evaluating \eqref{eq:split_crit} naively would require a high resolution sampling of the model in the node's volume, considering all pixel ray measurements that traverse it. However, we observe that, due to the particular structure of the inverse sensor model, we can make a conservative decision based on the measured depth $z$ and the point's query depth $r = \mathbf{T}_{WC_k}^{-1} {_W\mathbf{p}}\vert_z$ on whether to split the node by only considering the span of measurements $[z_{\min}, z_{\max}]$, and query depths $[r_{\min}, r_{\max}]$ in the node's volume, as illustrated in Fig. \ref{fig:voxels_3d}. 

While the span $r_{\min,\max}$ can be easily computed from the node's position and size, $z_{\min,\max}$ is more involved. To compute the latter, the node is projected into the depth image defining a bounding box (BB). More importantly, instead of evaluating all pixels within the BB, we sample a pre-computed \edit{set} of min-max pooling images.
}

Each pooling image $\mathbf{P}^f_k \in \mathbb{R}^{H \times W}$ with $f \in \{1, \dots, f_{\max}\}$ aggregates the information of the original depth image for different square areas $A^f = (2^f+1)^2$ centred around each pixel with coordinates $u$ and $v$. More specifically, pixel $\mathbf{P}^f_k[u,v]$ holds the span of measurement $z_{\min,\max}$ within $A^f$, as well as a validity and an image crossing state to handle invalid data and image boundaries. These pooling images are computed recursively from the previous level by summarising $\mathbf{P}^{f-1}_k[u\pm2^{f-2},v\pm2^{f-2}]$ as shown in Fig. \ref{fig:voxels_3d}. Once this structure is computed, a conservative evaluation of \eqref{eq:split_crit} can be made by simply computing a BB of the \edit{projected node} in the image and  querying the max-min pooling at the level with the maximum square size that is still fully contained in the BB, reducing the amount of queries needed to decide whether to split the node to no more than 4 in the majority of cases.
\edit{In summary, by considering measurement spans we reduce the problem of evaluating (\ref{eq:split_crit}) from a 3D sampling to a 2D one, which is further reduced to a few samples using a pre-computed set of pooling images. }

\subsection{Multi-resolution Probabilistic Occupancy Fusion}
\label{section:Fusion}
While data at node level may be at various resolutions, \edit{once voxel block level is reached, each block} is evaluated at a common resolution scale $s_c$, further improving performance, reducing aliasing effects and simplifying interpolation required by operations like raycasting and meshing.
Similar to \cite{Vespa2019}, measurements are integrated at a mip-mapped scale in the voxel block based on the current distance from the block centre to the camera. Moreover, we integrate blocks that are known to only contain frontiers from free to unknown space not finer than scale $s_f$. \edit{This way, the maximum level of detail ($s_f >= 0$) of the frontier boundaries can be chosen independently of the surface (see Fig. \ref{fig:quadtree} for $s_f = 1$}).

The desired integration scale $s_d$ may change with the distance.
In contrast to \cite{Vespa2019}, we apply a scale change hysteresis requiring the camera to move a certain amount. \edit{Given the focal length $c$ and the map resolution $v_{\text{res}}$, the camera has to move} $\Delta _C\mathbf{m}_z = 0.25 \times c \times v_{\text{res}}$ closer to or further away from the voxel block before changing the scale again, to avoid constant scale changes of blocks located at a scale boundary.

We adapt the propagation strategies applied in \cite{Vespa2019}, to our occupancy map representation to keep the hierarchy consistent between scale changes.
We replace the parent with the mean of all observed children when up-propagating information to a coarser mip-mapped scale ($s_d > s_c$). However, when down-propagating to a finer integration scale ($s_d < s_c$), we allocate the data first and assign the parent's value to all its children. In either case, we do not change scale immediately. Instead we wait for the block to be fully projected into the image plane multiple times and observed at the desired scale $s_d$. During this time we update the data at both scales $s_c$ and $s_d$. Once the changing condition is fulfilled we switch to the new integration scale, deleting the previous buffer. This reduces artefacts by smoothing values during initialisation and preventing blocks that are occluded or just about to exit the camera frustum from changing scale. \edit{In comparison to prior work our new down-propagation strategy enables eliminating all helper variables stored in each voxel, thereby drastically reducing memory consumption.}

Once all information from a depth image $\mathbf{D}_k$ is integrated into the map, the mean values of all updated blocks are up-propagated to the mip-mapped scale within each block to enable tri-linear interpolation between neighbours of different scales.
Additionally, the maximum occupancy of all updated blocks and nodes is propagated through all levels up to the root of the tree to support fast hierarchical free space queries. Finally, pruning is applied to merge node children whose weighted log-odd occupancy is close enough according to a lower threshold $L_{\min}$. This extra step simplifies the tree when the occupancies converge to similar values.

\subsection{Multi-resolution Ray-casting and Meshing Modules}
\label{subsec:raycasting}
Fast raycasting is required to render the surface in real-time and/or to enable tracking as part of a full dense SLAM system. With the up-propagated maximum occupancy and observed state stored in our data structure, we implemented a multi-resolution ray-casting strategy to quickly move a ray through large volumes of free space \cite{Knoll2006}.

A meshing module is also provided which computes an adaptive resolution dual mesh following \cite{Wald2020}. This approach had been adapted to support our two tier data structure.

\section{Evaluation}
\label{sec:experimental_results}
\begin{figure*}[htb]
    \centering
    \subfloat[]{\includegraphics[trim={0.0cm 3.3cm 0.0cm 3.3cm},clip,height=2.1cm]{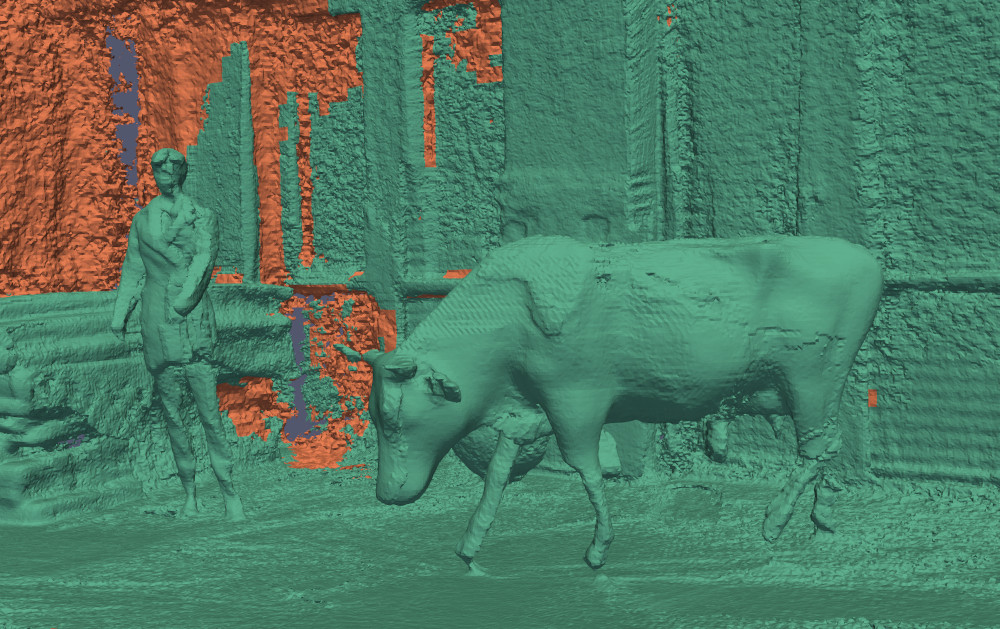}}
    \hspace{0.1cm}
    \subfloat[]{\includegraphics[trim={0.0cm 3.2cm 0.0cm 1.0cm},clip,height=2.1cm]{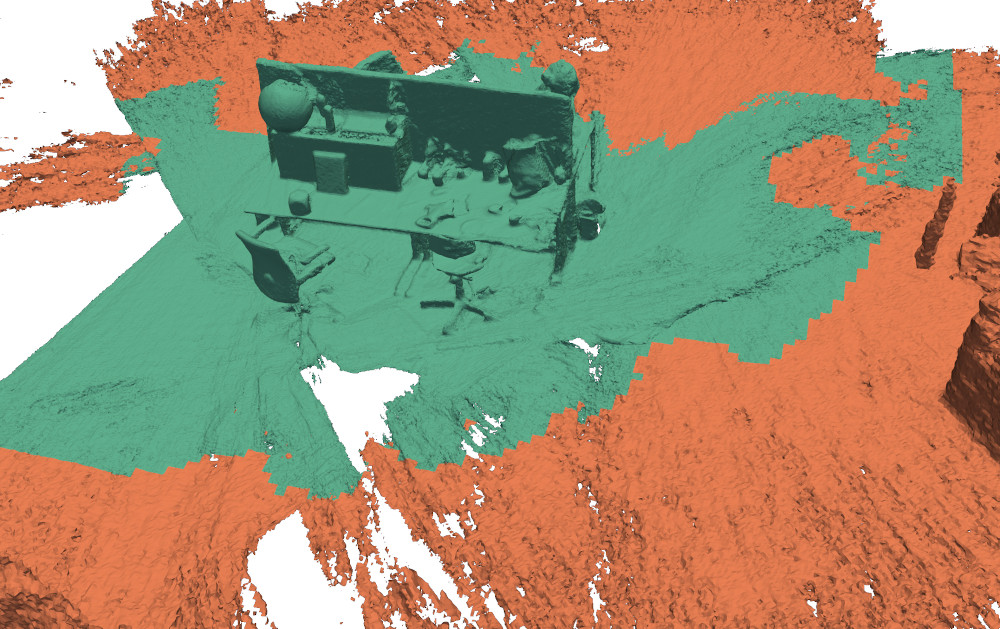}}
    \hspace{0.1cm}
    \subfloat[]{\includegraphics[trim={0.0cm 2.1cm 0.0cm 2.1cm},clip,height=2.1cm]{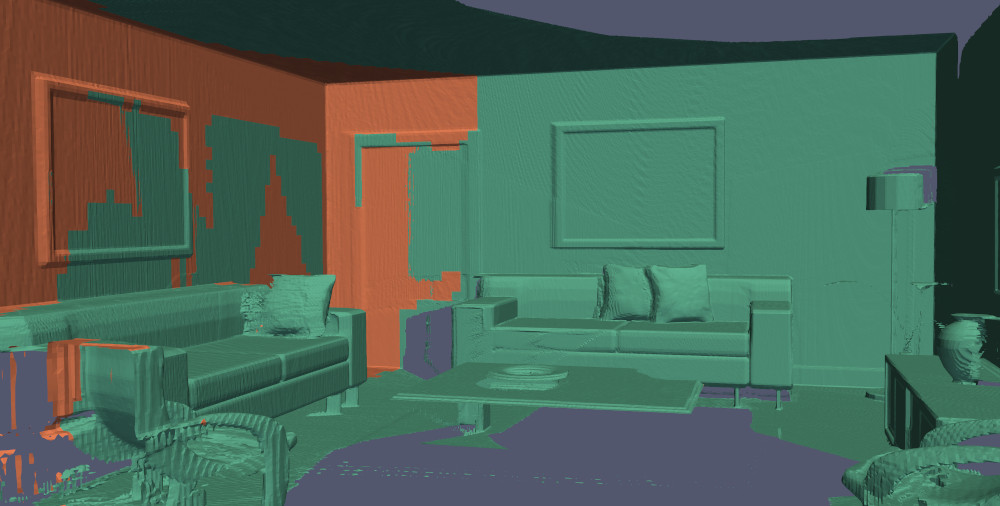}}
    \hspace{0.1cm}
    \subfloat[]{\includegraphics[height=2.1cm]{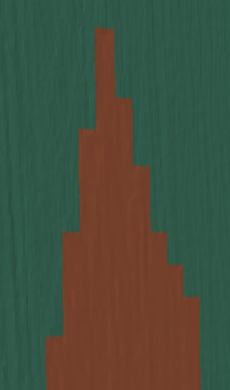}}
    \hspace{0.1cm}
    \subfloat[]{\includegraphics[height=2.1cm]{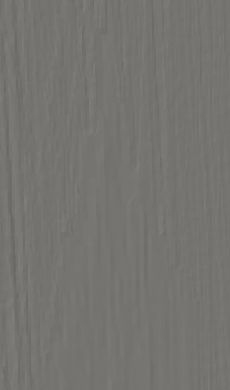}}
    \caption{Our multi-scale mesh reconstruction for different sensor types with green encoding sampling scale 0 (finest) and orange scale 1. (a) In the \textit{Cow and Lady} dataset and (b) \textit{TUM} dataset, scale is adapted to reflect higher Kinect sensor noise, particularly for places far from sensor trajectory, which loops around the central desk. (c) In the \textit{ICL-NUIM} synthetic dataset a similar model is used to show how, in lack of noise, no artefacts appear due to scale changes. (d) A close up in ICL showing an scale change in a wall. (e) Same as (d) but with uniform color, showing the lack of artefacts in the transition.}
    \label{fig:mesh}
\end{figure*}
This section presents our experimental results.
All our tests were performed on an Intel Core i7-8750H
CPU operating at 2.20 GHz, 16 GB of memory, and running Ubuntu 18.04.
We used GCC 7.4.0 with OpenMP acceleration for software compilation.
For computing the TSDF with \textit{voxblox} \cite{Oleynikova2017} the \textit{fast} method was used in all comparisons, \textit{OctoMap} \cite{Hornung2013}, \textit{UFOMap} \cite{Duberg2020} and \textit{voxblox} were run in ROS.
The parameter values used in the experiments are shown in Table \ref{tab:params} unless otherwise specified.
Three widely known datasets were used to quantitatively evaluate our system: \textit{TUM} RGB-D \cite{Sturm2012},  \textit{Cow and Lady} \cite{Oleynikova2017} and \textit{ICL-NUIM} \cite{Handa2014}.

\begin{table}[htb]
    \centering
    \begin{tabular*}{\columnwidth}{@{\extracolsep{\fill} } c c c c}
            \hline
		$f_{\max} = 5$ (\text{QVGA}) & \multicolumn{3}{c}{$s_f = 0$ ($v_{\text{res}} = 8\text{ cm}$)}
		 \\
		
		$f_{\max} = 6$ (\text{VGA})&
		\multicolumn{3}{c}{$s_f = 1$ ($v_{\text{res}} \in \{1 \text{ cm}, 2 \text{ cm}\}$)} \\
        \hline
		$l_{\min, \text{total}} = -100$ &
		$l_{\min, \text{iter}} = -5.015$ &
		\multicolumn{2}{c}{$L_{\text{min}}=0.95 \times l_{\min, \text{total}}$} \\
		
		$\sigma_{\min} = v_{\text{res}}$ &
		$\sigma_{\max} = 3 \times v_{\text{res}}$  &
		$z_{\mathrm{np}} = 0.4$ m & $z_{\mathrm{fp}} = 6$ m \\
		$\tau_{\min} =  3 \times v_{\text{res}}$ &
		$\tau_{\max} = 12 \times v_{\text{res}}$ &
		\multicolumn{2}{c}{$w_{\max} = l_{\min,  \text{total}} / l_{\min,  \text{iter}}$} \\
		
        \hline
        Cow and Lady       & \multirow{2}{*}{$\sigma(z) = 0.0025 \, {z}^2$} & \multicolumn{2}{c}{\multirow{2}{*}{$\tau(z) = 0.05 \, z$}} \\
        FR3 - Long Office  &  &\\
        ICL - LR2          & $\sigma(z) = 0.05 \, {z}$      & \multicolumn{2}{c}{$\tau(z) = 0.05 \, z$} \\
        \hline
    \end{tabular*}
    \caption{Parameter values used in the experiments.}
    \label{tab:params}
\end{table}

\subsection{Reconstruction Accuracy}
To evaluate how our method performs in terms of surface reconstruction, we evaluate in the \textit{ICL-NUIM Living Room 2} dataset  \cite{Handa2014}. 
Table \ref{tbl:mesh} shows the reconstruction \ac{RMSE} to the ground truth mesh as computed by the \textit{SurfReg} tool provided by the dataset, with given poses and without extra ICP alignment.
Our method outperforms previous approaches on both tested resolutions.

Additionally, we investigate the potential degradation induced by down-sampling the input image. It can be seen the effect on the reconstruction metric in all cases is minor, motivating the use of this later step for improving both running time and memory usage.
To show how our method can adaptively select sampling resolution, multiple reconstructions are presented in Fig. \ref{fig:mesh} with color encoding scale. 

\begin{table}[htb]
    \centering
    \begin{tabular*}{\columnwidth}[b]{@{\extracolsep{\fill} } l c c c}
        & Image scale & 1 cm & 2 cm \\
        \hline
        \multirow{2}{*}{Ours} & $\times1$   & 0.0146 & \textbf{0.0193} \\
                              & $\times0.5$ & \textbf{0.0250}          & \textbf{0.0341} \\
        \hline
        \multirow{2}{*}{SE OFusion} & $\times1$   & 0.0166        & 0.0259 \\
                                    & $\times0.5$ & 0.0478        & 0.0465 \\
        \hline
        \multirow{2}{*}{SE TSDF} & $\times1$    & \textbf{0.0142} & 0.0200 \\
                                 & $\times0.5$  & 0.0429           & 0.0367 \\
        \midrule
    \end{tabular*}
    \caption{Reconstructed mesh RMSE [m] on the \textit{ICL-NUIM LR2} dataset for different resolution and image subsampling.}
    \label{tbl:mesh}
\end{table}
\begin{figure}[htb]
    \centering
    \subfloat{\includegraphics[trim={0.0cm 7.0cm 0.0cm 2.0cm},clip,width=0.49\columnwidth]{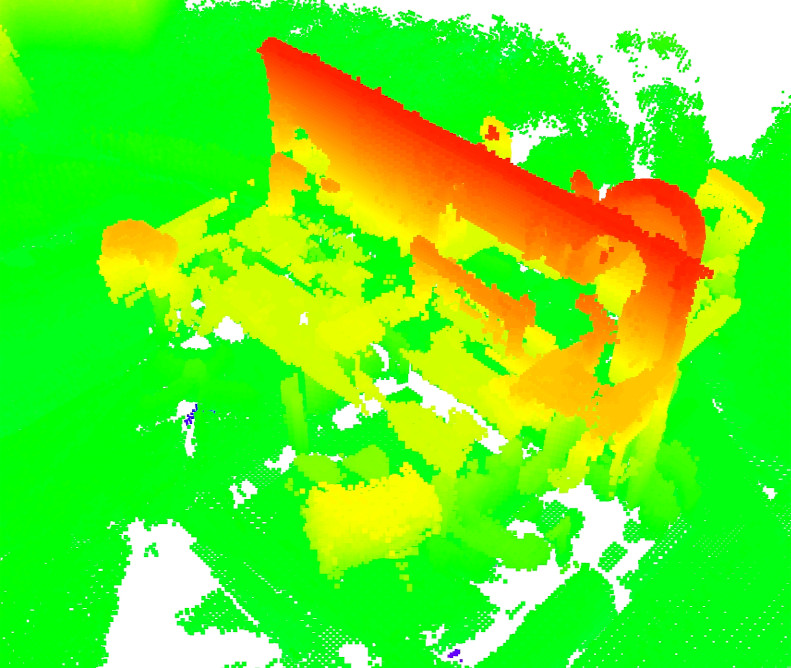}}
    \hspace{0.005cm}
    \subfloat{\includegraphics[trim={0.0cm 7.0cm 0.0cm 2.0cm},clip,width=0.49\columnwidth]{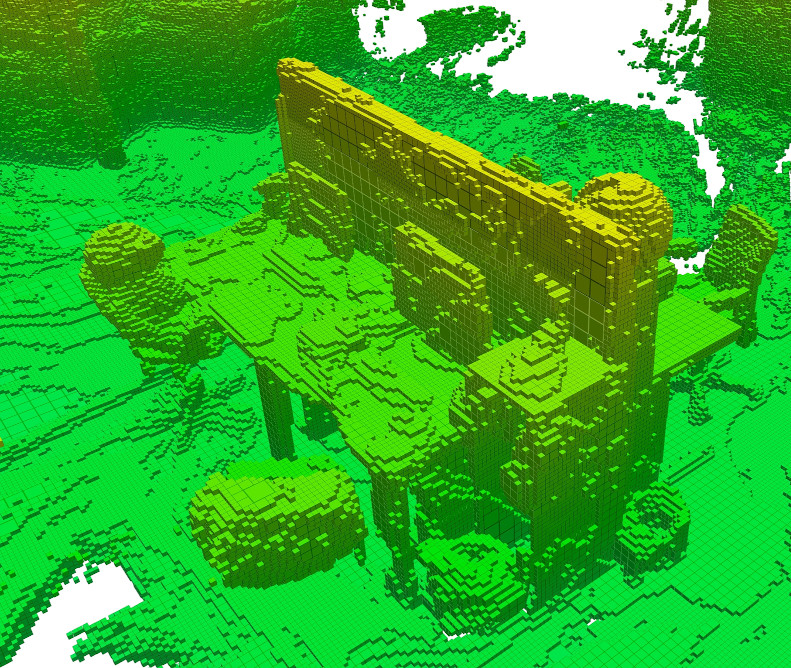}}
    \caption{Voxelised reconstruction in FR3 - Long Office for \textit{UFOMap} ($n = 0, d \hat{=} 16cm$, their visualiser) (\textbf{left}) and our system (\textbf{right}). Note the holes in the \textit{UFOMap} surface due to non-conservative free space allocation.}
     \label{fig:long_office_compare}
\end{figure}

\subsection{Runtime Performance}
\begin{figure*}[htb]
    \centering
    \subfloat{\includegraphics[trim={5.0cm 3.5cm 10cm 5cm},clip,width=0.33\textwidth]{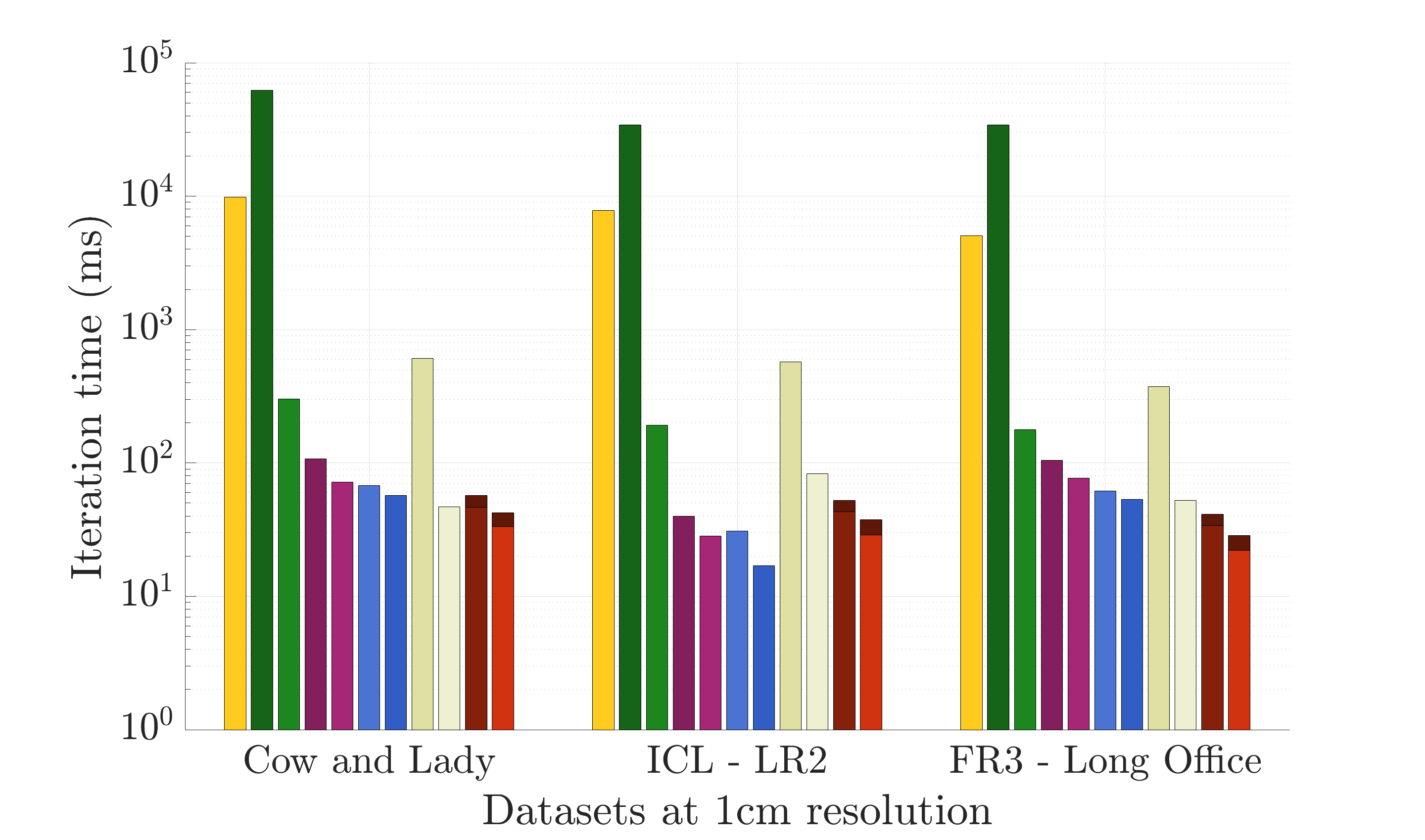}}
    \subfloat{\includegraphics[trim={5.0cm 3.5cm 10cm 5cm},clip,width=0.33\textwidth]{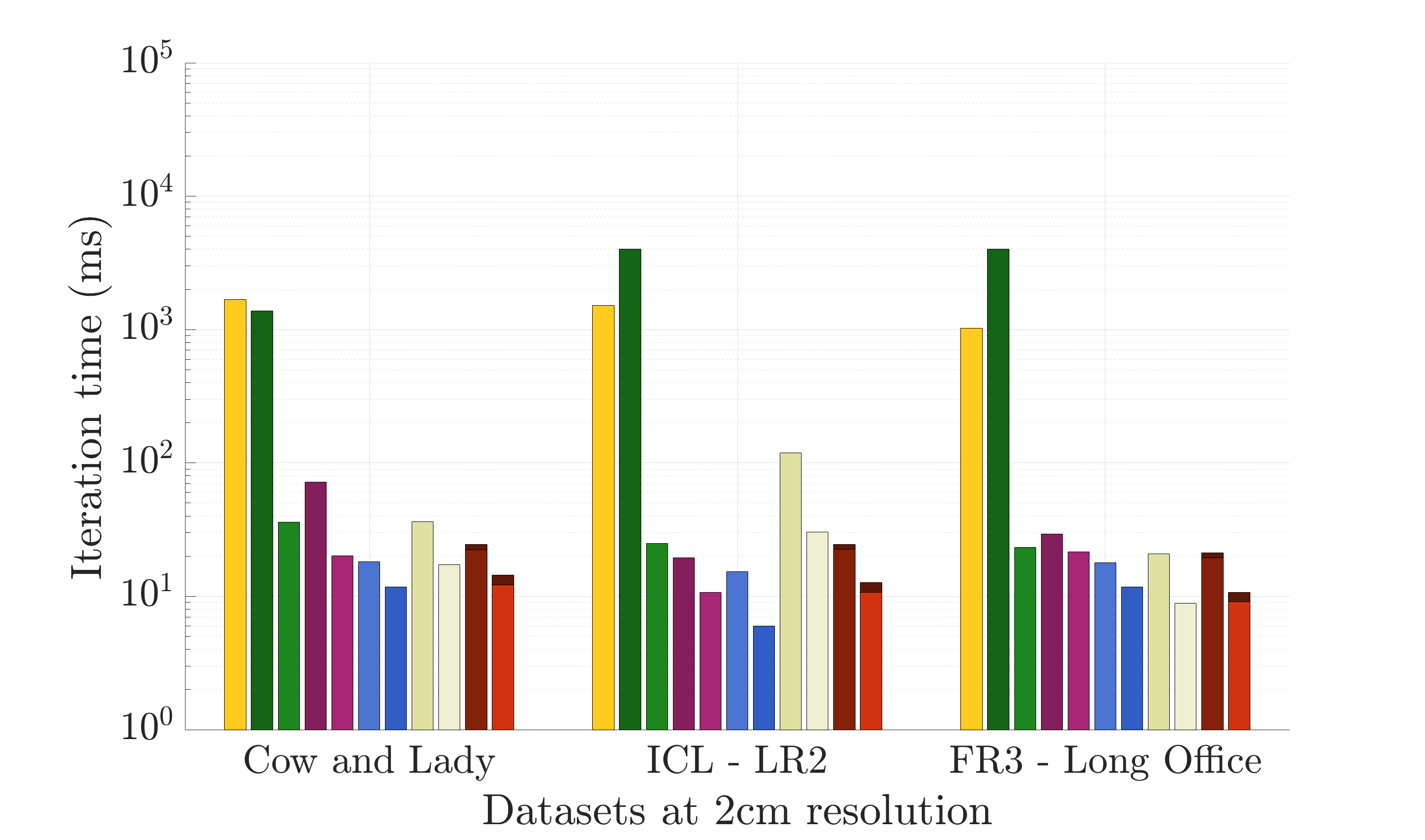}}
    \subfloat{\includegraphics[trim={5.0cm 3.5cm 10cm 5cm},clip,width=0.33\textwidth]{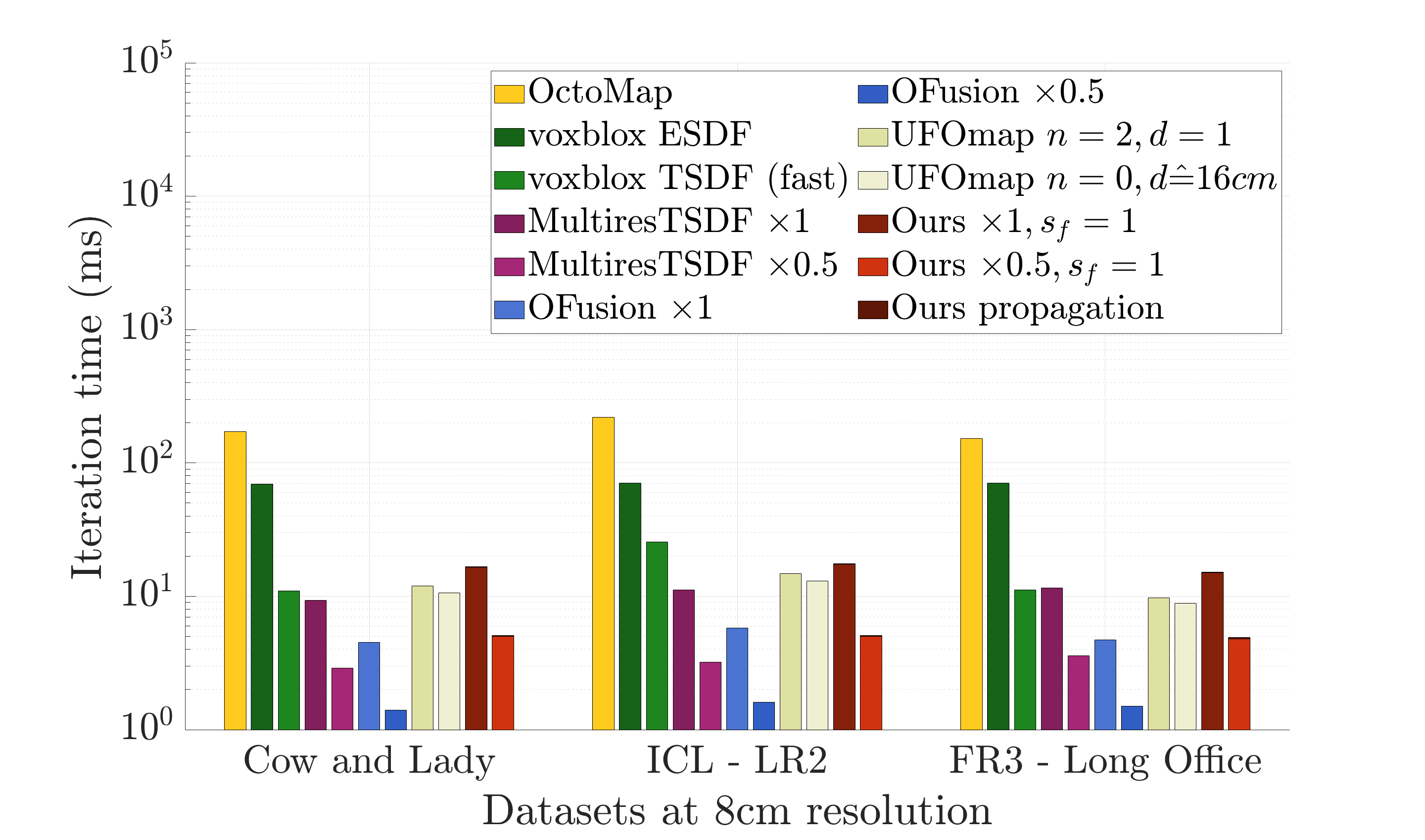}}
    \caption{Timings for our system alongside other planning libraries, including the widely used \textit{OctoMap} \cite{Hornung2013}, for 1cm (\textbf{left}), 2cm (\textbf{centre}) and 8cm (\textbf{right}) maximum resolution. Notice the logarithmic scale. In the case of \textit{voxblox} both timings for TSDF and ESDF calculation are shown, as both are needed for path planning. For some libraries including ours, results are presented for full resolution input images ($\times 1$) and down-sampled images ($\times 0.5$). The cost for computing data up-propagation (pooling) in our method is also separated to highlight the cost of having this multi-resolution feature in our map.}
    \label{fig:timings}
\end{figure*}

Running times for our system and competing methods can be seen in Fig. \ref{fig:timings}. Our method performs best at 1cm resolution for both image configurations, beating other methods in datasets using real sensors. 
\edit{We attribute this mainly to our adaptive-scale approach which leverages the additional computations needed to accurately and conservatively represent free-unknown space with economising resolution where not needed. 
As maximum resolution decreases to 8cm, or in small datasets, e.g. ICL-NUIM, the benefits of adaptive-scale are less pronounced while small costs like computing a pooling image become relatively larger. This is expected and consistent with the high-resolution/large-scale environments targeted by the algorithm.}

\edit{Methods like \textit{OFusion} and \textit{UFOMap} (\textit{fast}) also achieve excellent performance but at the expense of mapping quality, as shown in Section~\ref{subsec:planning} and Fig.~\ref{fig:long_office_compare}.  Additionally, methods such as} \textit{voxblox} present reasonable results, but have the drawback of requiring an extra step of computing an  ESDF in order to allow direct planning on them, which becomes prohibitively expensive in high resolution cases. This limits its usability in cases requiring a high resolution navigable map in real-time. Finally \textit{OctoMap} presents significantly higher integration times than newer methods, since it targets low resolution LIDAR made maps.

\subsection{Multi-resolution, Memory and Efficiency}
To assess memory efficiency of our multi-resolution approach,
we compare our memory usage to supereight OFusion \cite{Vespa2018} and supereight MutiresTSDF \cite{Vespa2019}. In particular OFusion uses an octree, similarly to our system for storing both free and occupied space, while MultiresTSDF also utilises an octree but for sparsely allocating a narrow-band TSDF. 

As a conservative measurement of memory usage, we use the main process' Resident Set Size (RSS) as reported by the Linux kernel to compute the RAM usage.

\begin{table}[htb]
    \centering
    \begin{tabular*}{\columnwidth}[b]{@{\extracolsep{\fill} } l c c | c c | c c}
        & \multicolumn{2}{c}{Cow \& Lady} & \multicolumn{2}{c}{FR3 Long Office} & \multicolumn{2}{c}{ICL LR2} \\
        \cline{2-7}
        & 1cm & 8cm & 1cm & 8cm & 1cm & 8cm \\
        \hline
        Ours       & \textbf{1100}  & 65          & \textbf{493} & 62          & 397          & 105\\
        \hline
        SE OFusion & 1734           & \textbf{57} & 1009         & \textbf{56} & \textbf{192} & \textbf{54}\\
        \hline
        SE MultiresTSDF    & 2922   & 65          & 1728         & 68          & 406          & 56\\
        \midrule
    \end{tabular*}
    \caption{Memory consumption in MB at different voxel resolutions.}
    \label{tbl:memory}
\end{table}

Results can be seen in Table~\ref{tbl:memory}. For the intended use case of high resolution reconstruction and planning, our system overcomes competing methods in real life scenarios, mainly thanks to the adaptive sampling resolution at voxel block level. As resolution decreases, or in small datasets, e.g. \textit{ICL-NUIM}, where the relative voxel block size grows in a way that brings the allocation near to dense, the cost of storing this extra multi-resolution information along with our conservative allocation of frustum boundaries, takes a toll on the usage when compared to simpler methods like OFusion. This extra cost is justified by superior planning performance.

\subsection{Tracking Performance}

\label{subsec:tracking}
In this section, we evaluate our mapping system integrated to a Dense SLAM Pipeline similar to \cite{Newcombe2011}, which is included as an additional component of the presented library. 

Table \ref{tab:ate} shows trajectory accuracy results using the \ac{ATE} metric against TSDF and OFusion \cite{Vespa2018} methods on the FR1 - Desk and FR3 - Long Office sequences of the \textit{TUM} RGB-D datasets.
The same ICP tracking approach presented in \cite{Newcombe2011} is used for all pipelines. For point-cloud extraction our system relies on the efficient multi-resolution raycasting method described in Section \ref{subsec:raycasting}.
Results indicate that maps produced by our approach are suitable for accurate ICP tracking, with the \ac{ATE} being similar to that of previous methods.
\begin{table}[htb]
    \centering
    \scriptsize
    \begin{tabular*}{\columnwidth}{@{\extracolsep{\fill} } l c c c c}
        & \multicolumn{4}{c}{ATE (m)} \\
        \cline{2-5}
        & \multicolumn{2}{c}{FR1 - Desk} & \multicolumn{2}{c}{FR3 - Long Office} \\
        \cline{2-3} \cline{4-5}
        Pipeline & 1 cm & 2 cm & 1 cm & 2 cm \\
        \hline
        Ours               & 0.104 & 0.098          & 0.194           & 0.185 \\
        SE TSDF            & \textbf{0.099}          & 0.103          & 0.314           & FAIL \\
        SE OFusion         & 0.100          & \textbf{0.086} & \textbf{0.165}  & \textbf{0.172} \\
        \midrule
    \end{tabular*}
    \caption{\ac{ATE} of various pipelines on \textit{TUM} RGB-D datasets.
    }
    \label{tab:ate}
\end{table}

\begin{figure}[htb]
    \centering
    \subfloat{\includegraphics[trim=4cm 0cm 2.1cm 0cm,clip,width=0.5\columnwidth]{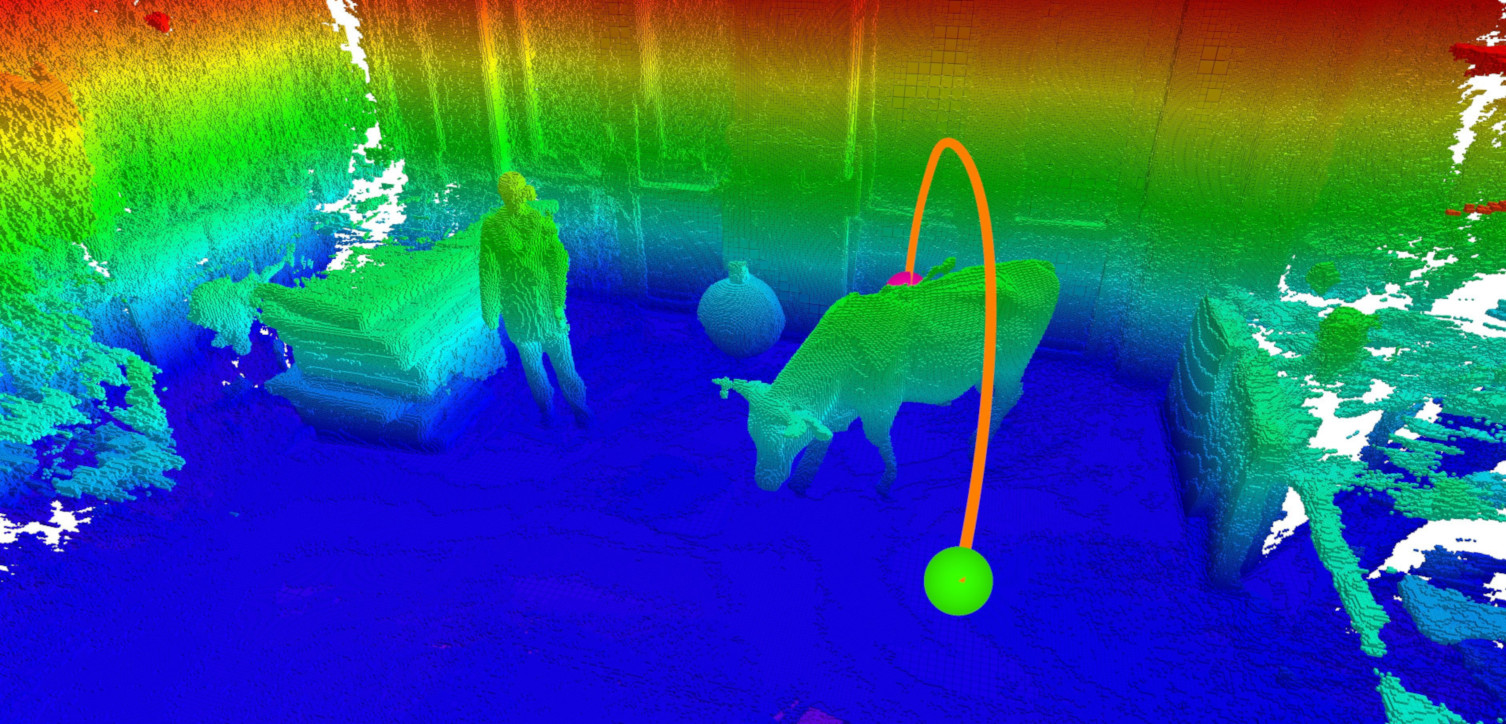}}
    \hspace{0.1cm}
    \subfloat{\includegraphics[width=0.44\columnwidth]{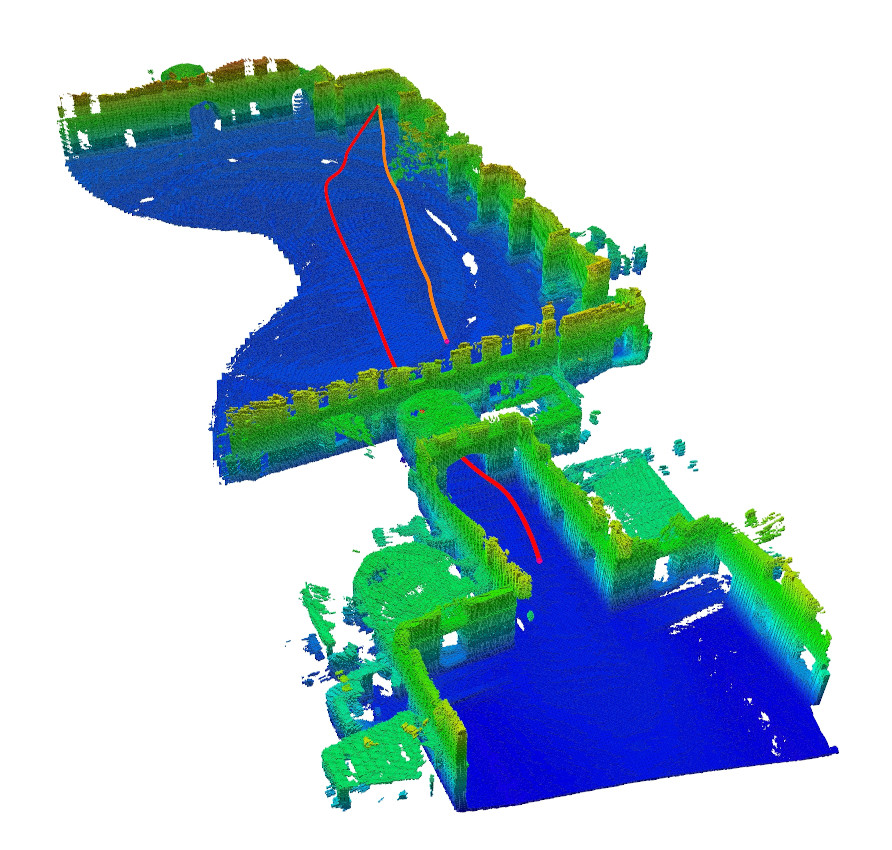}}
    \caption{(\textbf{left}) Planning in the \textit{Cow and Lady} dataset : Trajectory 1 over the cow obtained with our mapping system, OFusion is unable to solve this planning problem due to clutter. (\textbf{right}) Planning in The Newer College dataset \cite{Ramezani2020}:  Trajectores 3 (\textbf{orange}) and  4 (\textbf{red}). Remarkably the method is able to find the shortest path though the centre corridor.}
     \label{fig:traj1}
\end{figure}

\subsection{Planning}
\label{subsec:planning}
Finally, we evaluate our mapping system as input for path planning. We show that kinodynamically feasible and collision free quadrotor trajectories with map resolutions up to 1cm can be planned in real-time. To guarantee feasibility, we compute a \ac{SFC} from the start to the end position and optimise $10^{th}$ order Bernstein polynomial motion primitives within each segment. As a corridor primitive we use cylinders connected by spheres with a minimum radius $r_{\min}$. We use the open motion planning library's (OMPL \cite{Sucan2012}) informed rapidly-exploring random tree* (informed RRT* \cite{Gammell2014}) planner to create the \ac{SFC} connecting the start and end positions. OMPL confirms the safety of each corridor segment by verifying that none of the cylinder and sphere volumes is occupied. This is challenging using a regular volumetric grid where the number of checks grows cubically with the map resolution.

With our multi-resolution maximum occupancy queries, we utilise a `coarse-to-fine' collision checking approach recursively increasing the resolution in parts of the corridor where needed to reduce the number of checks required.
\begin{figure}[htb]
    \centering
    \subfloat{\includegraphics[width=0.48\columnwidth]{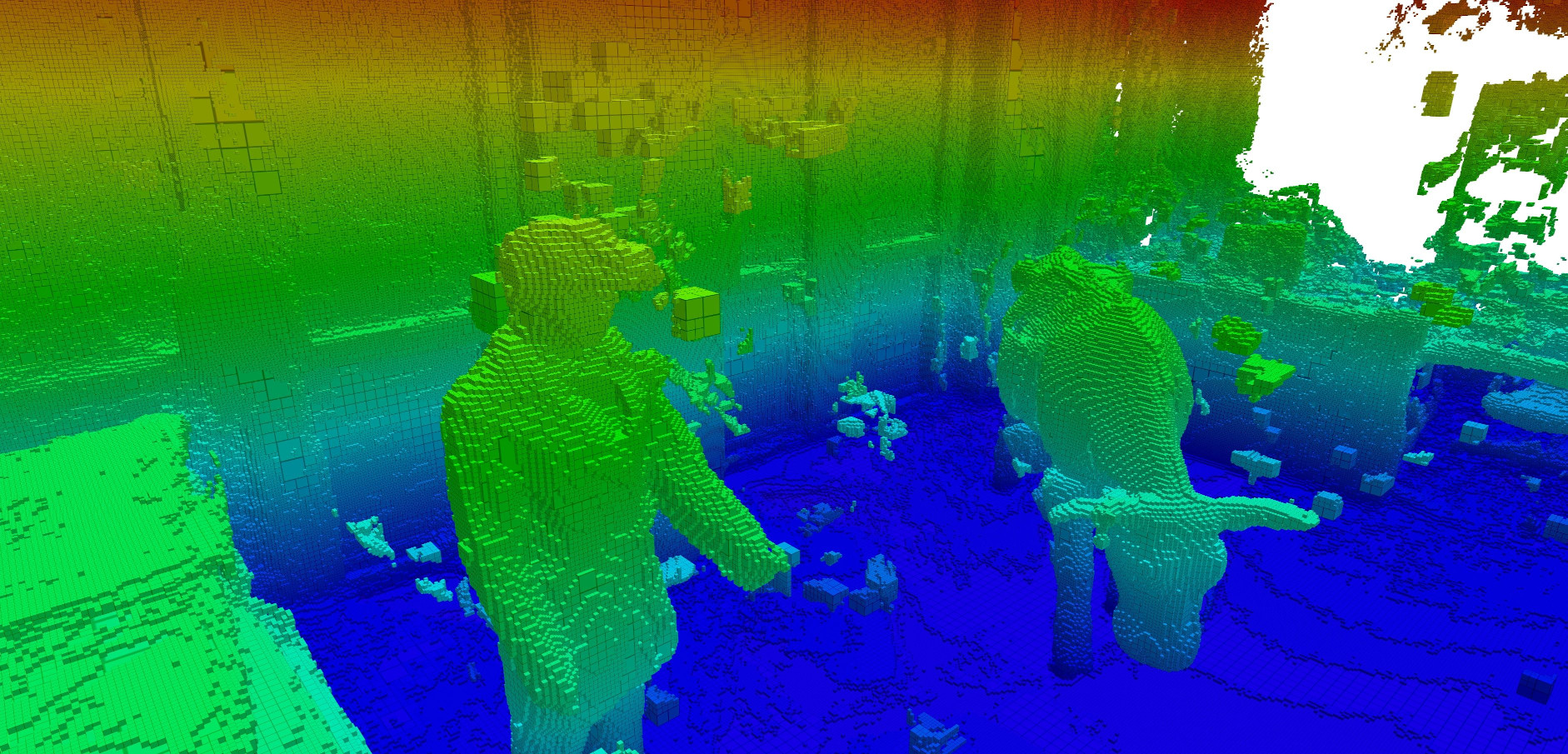}}
    \hspace{0.1cm}
    \subfloat{\includegraphics[width=0.48\columnwidth]{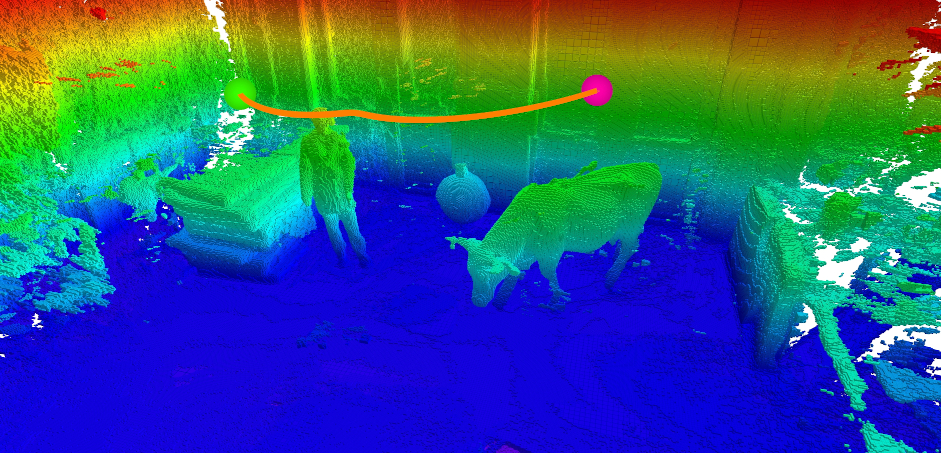}}
    \caption{(\textbf{left}) Clutter in OFusion makes the planner fail in several cases; (\textbf{right}) For a different problem (Trajectory 2) both systems find a solution but OFusion struggles with noise while planning times with our pipeline are significantly lower.}
    \label{fig:traj2}
\end{figure}

We investigate the use of our method for several path planning problems illustrated in Figs.~\ref{fig:traj1} and~\ref{fig:traj2}  and compare against the OFusion library, which also achieves excellent mapping times and provides some multi-resolution output. While our method can find suitable trajectories in every iteration, notably in cluttered datasets like the \textit{Cow and Lady} or large scale like \textit{The Newer College} \cite{Ramezani2020}, OFusion fails in several cases. This is because OFusion lacks a proper mechanism for leveraging information from different scales, instead relying on a simple rule of giving preemptiveness to data in high resolution voxel blocks over values stored at node levels. Given that the latter normally stores information about free space observations and the former allocates surface measurements, OFusion is fast but heavily biased against noise measurements and outliers. In our system we solve this issue by not only propagating data down from nodes to voxel blocks, but also by carefully considering the variation of occupancy inside the node's whole volume (not only the centre value), as described in Section~\ref{subsec:map_to_cam}, to avoid having the opposite effect of biasing towards free space.

To further confirm these aspects and highlight the benefits of our library, we modified OFusion to perform dense allocation, i.e. everything integrated at a single lowest voxel block level. The results in Table \ref{tbl:planning_timings} show the minimum solving time required by the planner to find a \ac{SFC}. While the dense allocation approach can remove OFusion noise and find suitable trajectories, the planning time required is much higher. This is easily explained by the lack of multi-resolution sampling capabilities, a key feature of our system.

\begin{table}[htb]
    \centering
    \begin{tabular*}{\columnwidth}[b]{@{\extracolsep{\fill} } l c c c c}
        & Traj 1 &  Traj 2 & Traj 3 & Traj 4\\
        \hline
        Ours                & \textbf{0.01} & \textbf{0.1} & \textbf{0.06} & \textbf{0.4} \\
        \hline
        SE OFusion          & 4             & FAIL (too noisy)            & 3             & FAIL (too noisy) \\
        \hline
        SE Dense            & 5             & 15           & 3             & 6 \\
        \midrule
    \end{tabular*}
    \caption{Minimum times (sec) to compute 'Safe Flight Corridor'. }
    \label{tbl:planning_timings}
\end{table}

\section{Conclusion} 
\label{sec:conclusion}
We have introduced a multi-resolution 3D mapping framework that is using an underlying two-tier octree data structure to encode log-odds occupancy probabilities. 
Thanks to explicit free space encoding in a hierarchical way, the approach supports fast collision checking, crucial in robotic path planning and collision avoidance, while providing high resolution reconstructions simultaneously.

Our framework was evaluated extensively in synthetic and real-world RGB-D datasets: we showed that surface accuracy is competitive with state-of-the-art TSDF-based frameworks, while enabling real-time or near-real-time operation even at centimetre-level resolutions. 
We finally show our maps used in different real-time 3D trajectory scenarios, including large scale LIDAR ones, to dramatically improve planning time without converting the map into ESDF as required by many existing approaches; thus providing seamless, unprecedented integration between mapping and planning.


\bibliographystyle{IEEEtran}
\bibliography{references/2021-icra-funk.bib}

\end{document}